\documentclass[journal]{IEEEtran}
\usepackage{cite}

\usepackage[pdftex]{graphicx}
\graphicspath{{../figs/}}
\DeclareGraphicsExtensions{.pdf,.jpeg,.png}
\ifCLASSINFOpdf
\else
\fi
\usepackage{amsmath}
\usepackage{physics}
\usepackage{siunitx}
\usepackage{mathtools}
\usepackage{amssymb}
\usepackage{gensymb}
\usepackage{multirow}
\usepackage{hyperref}
\usepackage{caption}
\usepackage{subcaption}
\usepackage{algorithmic, algorithm}
\begin{document}
%
\title{Adversarial Evaluation of Autonomous Vehicles in Lane-Change Scenarios 
}
%
%
%


\author{Baiming~Chen~\IEEEmembership{Student Member,~IEEE}, Xiang~Chen, Qiong Wu, Liang~Li*~\IEEEmembership{Senior Member,~IEEE}, %
\thanks{This work was supported in part by the National Key Research and Development Program of China under Grant 2018YFB0105101 and in part by Technological Innovation Project for New Energy and Intelligent Networked Automobile Industry of Anhui Province under Grant JAC2019030105. \textit{(Corresponding author: Liang Li.)}}%
\thanks{Baiming Chen, Xiang Chen and Liang Li are with the State Key Laboratory of Automotive Safety and Energy, Tsinghua University, Beijing 100084, China e-mail: cbm17@mails.tsinghua.edu.cn; tschenxiang@mail.tsinghua.edu.cn; liangl@tsinghua.edu.cn.}
\thanks{Qiong Wu is with Technology Center Anhui Jianghuai Automobile Co Ltd, Hefei 230601, China.}
}

\maketitle

\begin{abstract}

Autonomous vehicles must be comprehensively evaluated before deployed in cities and highways. However, most existing evaluation approaches for autonomous vehicles are static and lack adaptability, so they are usually inefficient in generating challenging scenarios for tested vehicles. In this paper, we propose an adaptive evaluation framework to efficiently evaluate autonomous vehicles in adversarial environments generated by deep reinforcement learning. Considering the multimodal nature of dangerous scenarios, we use ensemble models to represent different local optimums for diversity. We then utilize a nonparametric Bayesian method to cluster the adversarial policies. The proposed method is validated in a typical lane-change scenario that involves frequent interactions between the ego vehicle and the surrounding vehicles. Results show that the adversarial scenarios generated by our method significantly degrade the performance of the tested vehicles. We also illustrate different patterns of generated adversarial environments, which can be used to infer the weaknesses of the tested vehicles.
\end{abstract}


%
\IEEEpeerreviewmaketitle

\section{Introduction}
%
%
%
%

\begin{figure*}[ht]%
    \centering
    {{\includegraphics[width=1\linewidth]{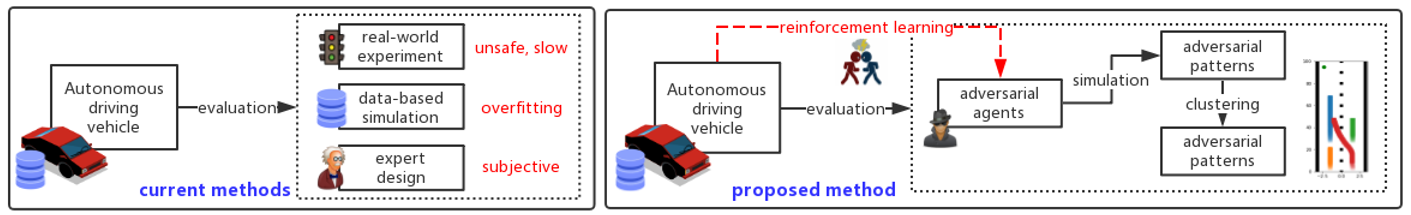} }}%
    \caption{A Comparison between the mainstream approaches and the proposed method for evaluation of autonomous vehicles. Most current mainstream methods are open-loop, which makes them inefficient. In contrast, the proposed framework is close-loop - the adversarial agents adaptively generate risky environments based on the behaviors of the tested vehicles. }
    \label{process}
\end{figure*}

\IEEEPARstart{A}{utonomous} vehicles are commonly believed to be a promising approach to decreasing traffic accidents in future transportation because they can avoid unreasonable behaviors of human drivers that could lead to fatal crashes~\cite{bagloee2016autonomous}. However, how to develop a safe autonomous driving system in complex environments is still an open problem~\cite{koopman2017autonomous}. 

One essential procedure to improve the safety of autonomous vehicles is to conduct a systematic evaluation before their deployment. The most popular approach in the industry is a data-based method called Naturalistic Field Operational Tests (N-FOT)~\cite{festa2008festa}. The principle of N-FOT is to test autonomous vehicles in naturalistic traffic environments, which are reconstructed by the driving data collected by sensor-equipped vehicles over a long time~\cite{aust2012evaluation}. 
One limitation of this method is inefficiency because of the rareness of risky scenarios in naturalistic environments. According to the National Highway Traffic Safety Administration (NHTSA), there were 6,734,000 police-reported motor vehicle traffic crashes and 33,654 fatal crashes in USA in 2018, while the total distances that the vehicles traveled was 3,237,500 million miles~\cite{national2018traffic}, which makes the average distance 0.48 million miles for each crash and 96.20 million miles for each fatal crash. The rareness of risky events makes the evaluation procedure extremely slow, even in simulation.

Another weakness of N-FOT is that it is usually static, which means that the distribution of the testing environments are fixed and can not evolve based on the behavior of the tested vehicles. In this case, the evaluation is not only inefficient but also incomplete - if some patterns of risky scenarios never appear in the dataset, they will be omitted in the evaluation procedure. In other words, the evaluation results heavily depend on the completeness of the traffic dataset, which is very difficult to verify~\cite{schelter2018automating}.

Recently, several approaches have been proposed to improve the efficiency of N-FOT. Zhao et al.~\cite{zhao2016accelerated} introduced the importance sampling technique with the cross-entropy method to accelerate the evaluation procedure in lane-change scenarios. While maintaining the accuracy, the evaluation is 2,000 to 20,000 times faster than the naturalistic driving tests in simulation. Feng et al.~\cite{feng2020testing} proposed an adaptive framework for testing scenario generation. From the Bayesian perspective, they iteratively update the distribution of risky scenarios based on the behavior of the tested vehicle. They also utilize an exploration strategy to simulate unseen scenarios. However, these methods still rely heavily on an initial dataset. Also, they focus on generating the initial conditions instead of modeling the time-sequential interactions between the ego vehicle and its surrounding vehicles.


Another promising direction to generate risky environments is via adversarial learning, which has been recently widely used for developing robust intelligence~\cite{li2019parallel}. For example, Pinto et al.~\cite{pinto2017robust} propose Robust Adversarial Reinforcement Learning (RARL) to train an optimal adversarial agent for modeling disturbances in zero-sum games and finally develop a robust controller. Bansal et al.~\cite{bansal2017emergent} suggest that sufficient complexity of the environment for training is required for a highly capable agent. With adversarial reinforcement learning and self-play, the agents learned a wide variety of complex and interesting skills in 3D physically simulated environments. However, directly applying existing adversarial learning algorithms for the evaluation of autonomous vehicles will be problematic. One issue is that in traffic scenarios, the ego vehicle and its surrounding vehicles are not fully competitive, which breaks the zero-sum assumption of the traffic interactions. Also, most current methods only search for the optimal adversarial environment, which is inefficient because a variety of risky patterns is desired to test the robustness of the self-driving agent. To our best knowledge, the adversarial learning framework has not been fully utilized for the evaluation of autonomous vehicles.

To address the above challenges, we build an adaptive framework in this paper to generate adversarial environments for tested vehicles with deep reinforcement learning. Specifically, we regard the environment vehicles as \textit{adversaries} that try to interfere with the tested ego vehicle, and train adversarial policies for their time-sequential decision-making. In this way, we aim to find the weaknesses of the tested vehicles and learn how to exploit them. To better model the mixed cooperative and competitive interactions between the ego vehicle and environment vehicles, we first design non-zero-sum reward functions based on domain knowledge and traffic rules and formulate the scenarios as Markov Decision Processes (MDPs). Then, during the training phase, we utilize ensemble reinforcement learning to collect local optimums of adversarial policies for the diversity of generation. After training, we cluster the generated risky scenarios with a nonparametric Bayesian approach since the number of potential local optimums is unknown. 
A comparison between current mainstream evaluation methods and the proposed method is shown in Fig.~\ref{process}.

The main contributions of this paper are summarized as follows. 
\begin{itemize}
    \item We build an efficient adaptive framework to generate time-sequential adversarial environments for the evaluation of autonomous vehicles.
    \item Non-zero-sum reward functions are designed to model the mixed cooperative and competitive interactions between the ego vehicle and environment vehicles.
    \item We utilize ensemble training to collect diverse risky scenarios and cluster them with a nonparametric Bayesian approach.
    \item A typical lane-change scenario is used for the evaluation of the proposed framework. Results show that adversarial environments significantly increase the collision rate for both rule-based and learning-based lane-change models. 
\end{itemize}

The rest of the paper is organized as follows. Section~\ref{section2} introduces the testing scenarios and terms used in the paper. Section~\ref{section3} demonstrates how to generate adversarial environments with ensemble reinforcement learning and cluster them with DP-Means. In Section~\ref{section4}, the adversarial environments are visualized, and two typical lane-change models are evaluated.

\section{Lane-Change Scenario}
\label{section2}
The lane-change scenario is used to show the benefit of the proposed adversarial evaluation method. 
We choose the lane-change scenario for several reasons. First, crash data from 2010 to 2017 shows that the sudden lane change caused about 17.0\% of total severe crashes, followed by speeding (12.8\%) and tailgating (11.2\%)~\cite{shawky2020factors}, which indicates that the lane change maneuver is one of the most serious causes of car accidents. Also, auto lane change is a frequent driving maneuver and is considered an important and challenging task for autonomous driving, which recently attracts lots of attention in both academia~\cite{nilsson2016lane} and industry~\cite{endsley2017autonomous}. In addition, it is an important part of microscopic traffic simulation and has a considerable effect on the analysis results of these models~\cite{mathew2014lane}. Given all the above reasons, we believe that the evaluation of the lane-change behavior of autonomous vehicles is an important topic.

\begin{figure}[b]
\centering
\includegraphics[width=3 in]{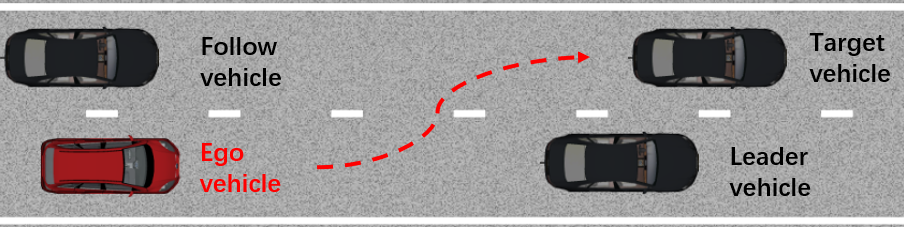}
\caption{A typical lane-change scenario. The ego vehicle is trying to make a left lane change in the presence of three surrounding vehicles.}
\label{lanechange}
\end{figure}

We use a typical lane-change scenario shown in Fig.~\ref{lanechange}. There are four vehicles in this scenario, including three surrounding vehicles (follow vehicle, leader vehicle, and target vehicle) and one ego vehicle that is trying to make a left lane change. Moreover, we simplify the environment by assuming that only the ego vehicle performs lateral movement for a lane change, while the other three environment vehicles go straight in the lane, which is a commonly used setting in the literature~\cite{shi2019driving,duan2020hierarchical}. The lane width is 3.2 meters, while the vehicles have a width of 1.85 meters and a length of 4.83 meters. These data will be later used for collision detection.


To generate realistic scenarios that mimic the real-world traffic, we use the guidance from a naturalistic database published by the Safety Pilot Model Deployment (SPMD) program~\cite{bezzina2014safety}. The SPMD program recorded naturalistic driving data in Ann Arbor, Michigan with 2,842 equipped vehicles for more than two years. The MobilEye cameras installed on the vehicles provide the relative distance between the ego vehicle and the front vehicle. The relative velocity is calculated based on temporal difference.

\begin{figure}[h]
\centering
\includegraphics[width=0.7\linewidth]{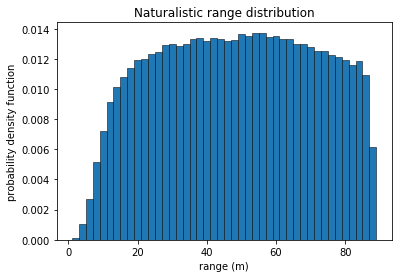}
\caption{The probability density function of naturalistic range based on the SPMD database.}
\label{fig:pdf}
\end{figure}

We use the SPMD database to help setting initial conditions for each run of experiment. First, the initial longitudinal distance between the ego vehicle and the leader vehicle $x_{leader}$ is sampled from the empirical distribution that calculated from the database (Fig.~\ref{fig:pdf}). Then, the longitudinal distance between the follow vehicle and the target vehicle $x_{target-follow}$ is sampled from the same distribution. Since the longitudinal distance between the ego vehicle and the follow vehicle $x_{follow}$ is not provided in the dataset, it is set to follow a Gaussian distribution: $x_{follow}\sim\mathcal{N}(\mu_x,\,\sigma^{2}_x)$, where $\mu_x$ is 0 $m$ and $\sigma_x$ is 5 m. The longitudinal distance between the ego vehicle and the target vehicle $x_{target}$ is then equal to $x_{follow} + x_{target-follow}$. The initial velocities of the four vehicles in the scenario $v$ are sampled from a Gaussian distribution: $v\sim \mathcal{N}(\mu_v,\,\sigma^{2}_v)$, where $\mu_v$ is 10~m/s and $\sigma_v$ is 4~m/s.


Based on the above initial settings, the lane-change experiment can be formulated as an episodic game. The ego vehicle is trying to make a left lane change within a limited distance $x_{lim}=300$~m and time $t_{lim}=30$~s. Referring to the testing protocols published by NHTSA~\cite{national2013lane}, the lane change is judged to be successful if the whole body of the ego vehicle is in the target left lane.

The initial longitudinal behaviors of the surrounding vehicles are controlled based on the Intelligent Driver Model (IDM)~\cite{treiber2000congested}. With IDM, the longitudinal acceleration of the vehicle $\alpha$ is
\[
\dot{v}_\alpha = \dv{v_\alpha}{t} = \alpha\left( 1- \left( \frac{v_\alpha}{v_0}\right)^\delta-\left(\frac{s^*\left(v_\alpha, \Delta v_\alpha\right)}{s_\alpha}\right)^2\right),
\]
where
\[
s^*\left(v_\alpha, \Delta v_\alpha\right)=s_0+v_\alpha T + \frac{v_\alpha\Delta v_\alpha}{2\sqrt{ab}}.
\]
The descriptions and values of parameters used for IDM in this paper are shown in Table~\ref{IDM}.

\begin{table}[t]
\renewcommand{\arraystretch}{1.3}
\caption{Parameters of IDM}
\label{IDM}
\centering
\begin{tabular}{c c c}
\hline
Parameter & Description & Value\\
\hline\hline
$v_0$ & Desired velocity & 10 m/s\\

$T$ & Safe time headway & 1.5 s\\

$a$ & Maximum acceleration & \SI[per-mode=symbol]{1}{\meter\per\second\squared}\\

$b$ & Comfortable Deceleration & \SI[per-mode=symbol]{1.67}{\meter\per\second\squared}\\

$\delta$ & Acceleration exponent & 4\\

$s_0$ & Minimum distance & 2 m\\
\hline
\end{tabular}
\end{table}

\section{Adversarial Evaluation Approach}
\label{section3}

We introduce the proposed framework of adversarial evaluation in this section. To evaluate autonomous driving agents, we first generate adversarial environments where the surrounding vehicles (adversaries) are trying to prevent the ego vehicle from finishing the task. Then, we cluster the generated environments and provide risky patterns for the tested ego vehicle. The proposed adversarial evaluation method can significantly accelerate and supplement traditional evaluation procedures.

\subsection{Adversarial Environment Generation}
\label{adv}
Autonomous driving systems are usually developed and evaluated based on naturalistic data. However, the database is always limited in two ways. First, the capacity of the database can never be infinitely large, which means it won't be able to cover every scenario that could happen in the real world, considering the time and budget. Second, risky events are rare in the database~\cite{zhao2016accelerated}. For these two reasons, the autonomous driving systems developed based on the database are likely to be overfitted, which makes them unpredictable in scenarios that never appear in the database. Thus, it is risky to test them in environments generated only by the data distribution $P_{data}$. In this section, we propose an adaptive method to efficiently generate adversarial environments $P_{adv}$ for the tested vehicle, aiming to supplement the missing parts of $P_{data}$.

We first formulate the lane-change scenario as a two-player Markov Game represented by a tuple $\left(\mathcal{S}, \mathcal{A}_1, \mathcal{A}_2, p, r_1, r_2, \gamma \right)$, where
\begin{itemize}
    \item $\mathcal{S}$ is the state space;
    \item $\mathcal{A}_1$ is the action space for the ego vehicle;
    \item $\mathcal{A}_2$ is the action space for the environment vehicles (adversaries);
    \item $p:\mathcal{S}\times \mathcal{A}_1 \times \mathcal{A}_2 \times \mathcal{S} \to \mathbb{R}$ is the state transition probability of the physical environment;
    \item $r_1:\mathcal{S}\times \mathcal{A}_1 \times \mathcal{A}_2 \to \mathbb{R}$ and $r_2:\mathcal{S}\times \mathcal{A}_1 \times \mathcal{A}_2 \to \mathbb{R}$ are the immediate rewards for the ego and surrounding vehicles.
\end{itemize}

In a Markov game, each agent $i$ aims to maximize its own total expected return $R_i=\sum_{t=0}^{T} \gamma^t r^t_i$ with a policy $\pi_i:\mathcal{S} \to \mathcal{A}_i$, where $T$ is the time horizon. 

From the perspective of adversaries, the ego vehicle can be regarded as a part of the environment, and then the Markov Game degrades to a Markov decision process (MDP). Here, we regard the multi-adversaries as one single agent, meaning they share the same reward function and are fully-cooperative to challenge the ego vehicle. This assumption is achievable in intelligent transportation systems thanks to V2X infrastructures. 

Reinforcement learning is a powerful tool to solve MDPs and find optimal or suboptimal policies for learning agents. Specifically, deep deterministic policy gradient (DDPG) has been widely used to solve MDPs with continuous action space~\cite{lillicrap2015continuous}. It has also been used to develop autonomous driving agents by many researchers~\cite{qiao2018automatically,huang2019autonomous}. Thus, in this paper, we also use DDPG to train adversaries and generate risky scenarios. However, other algorithms for continuous control tasks~\cite{schulman2017proximal,haarnoja2018soft} are also directly applicable to our framework.

DDPG is a reinforcement learning algorithm with an actor-critic architecture. The actor $\mu(s|\theta^\mu)$ is a parameterized function that specifies the current policy which deterministically maps states to a specified action. The critic $Q(s,a)$ is the action-value function which describes the expected return after taking an action $a_t$ in state $s_t$ an following the policy $\mu$ afterwards:
\[
Q^\mu\left(s_t,a_t\right)=\mathbb{E}_{r_{i\geq t},s_{i\geq t}\sim E,a_{i\geq t}\sim \mu }\left[R_t|s^t, a^t\right].
\]
The critic is updated following the Q-learning~\cite{watkins1992q} which is based on Bellman equation. Consider the function approximator parameterized by $\theta^Q$, the critic is optimized by minimizing the loss:

\[
\mathcal{L}\left(\theta^Q\right)=\mathbb{E}_{s_t\sim \rho^\beta,a_t\sim \beta,r_t \sim E}\left[\left(Q^*\left(s_t,a_t|\theta^Q\right)-y_t\right)^2\right],
\]
where
\[
y_t=r\left(s_t,a_t\right)+\gamma Q\left(s_{t+1},\mu\left(s_{t+1}\right)|\theta^Q\right),
\]
where $\beta$ is different behavior policy and $\rho$ represents the state distribution. This indicates that Q-learning is an off-policy algorithm. Thus, experience replay buffer can be used to eliminate the time correlation and improve the sample efficiency~\cite{mnih2015human}. 

The actor is updated by following the policy gradient~\cite{silver2014deterministic}:
\[
\nabla_\theta^\mu J \approx \mathbb{E}_{s_t \sim \rho^\beta} \left[\nabla_a Q\left(s,a|\theta^Q\right)|_{s=s_t, a=\mu\left(s_t\right)} \nabla_\theta^\mu \mu\left(s|\theta^\mu\right)|_{s=s_t} \right].
\]

To make the update iterations stable, a copy of the actor and the critic network is created: $\mu^\prime(s|\theta^{\mu^\prime})$ and $Q^\prime(s,a|\theta^ {Q^\prime})$. The parameters of these target networks are slowly updated to track the learned models: $\theta^\prime = \tau \theta + (1- \tau )\theta^\prime$.

The reward function $r(s_t,a_t)$ is an essential part and directly determines how the adversarial vehicles behave. Previous studies in adversarial learning usually assume the game is zero-sum~\cite{pinto2017robust,bansal2017emergent}, where the reward functions of the adversaries and the ego agent are opposite:
\[
r_{adv} = -r_{ego}.
\]
Here, $r_{ego}$ is the driving performance of the ego vehicle. However, directly applying this reward function to the adversaries will lead to unreasonable behaviors. For example, the surrounding vehicles may learn to directly rush to the ego vehicle and create a crash, just to decrease the driving performance of the ego vehicle. These kinds of unreasonable scenarios are not very informative for autonomous vehicle evaluation since they are not likely to happen frequently in the real world. Thus, to generate more reasonable risky scenarios, we relax the zero-sum assumption and add another term to the adversarial reward function:
\[
r_{adv} = -r_{ego} + \beta r_{rule},
\]
where $r_{rule}$ is a penalty for violation of traffic rules, which prevents the adversaries from being irrational, and $\beta$ is a hyperparameter, which determines the rationality of the environment vehicles. With this augmented reward function, the adversarial vehicles in the environment will try to interfere with the ego vehicle while obeying the traffic rules. The concrete settings of the reward functions will be introduced in the experiment part.




Most current reinforcement learning algorithms aim to find the globally optimal policy for the learning agent. To avoid being stuck at local optimums, they usually spend huge computing effort for exploration~\cite{lillicrap2015continuous,mnih2015human}. However, this principle is inefficient for generating diverse adversarial scenarios. For the evaluation of autonomous vehicles, we prefer to collect diverse local optimums since they may represent various weaknesses of the tested vehicle. Thus, instead of avoiding the local optimums, we embrace them in this paper for efficiency and diversity. We propose the ensemble DDPG for local optimums (Algorithm~\ref{ensemble}). Instead of training one agent, we train $N$ agents with random initializations of the actor and the critic. Exploration is omitted for fast convergence to local optimums. For each agent, we stop training if a local optimum has been reached, or the cumulative reward of one episode $\Sigma_{t=1}^T\gamma^tr_t$ has reached some boundary $c$, which indicates that a challenging environment has been found for the tested ego vehicle.

\begin{algorithm}[t]
\caption{Ensemble DDPG for local optimums}
\begin{algorithmic}
\label{ensemble}
\FOR {adversarial agent} 
\STATE Randomly initialize the actor $\mu(s|\theta^{\mu})$ and the critic $Q(s,a|\theta^{Q})$
\STATE Copy for the target networks $\mu^\prime(s|\theta^{\mu^\prime})$ and $Q^\prime(s,a|\theta^ {Q^\prime})$
\STATE Initialize replay memory $\mathcal{D}$
\WHILE {\textbf{not} \textit{Converged} \OR $\Sigma_{t=1}^{T}\gamma^tr_t \geq c$}
\STATE Update $\theta^{\mu}$, $\theta^{Q}$, $\theta^{\mu^\prime}$and $\theta^{Q^\prime}$ with the DDPG algorithm without exploration
\ENDWHILE
\STATE Save $\mu(s|\theta^{\mu})$
\ENDFOR
\end{algorithmic}
\end{algorithm}

\subsection{Environment Clustering}
After training, each bootstrap from the ensemble can represent one kind of risky scenarios for the tested vehicle. However, some of them may represent similar scenarios. To better extract different types of adversaries from the ensemble, we propose to apply an unsupervised clustering of the training results.  

Direct clustering of the learned adversarial policies \{$\mu(s|\theta^{\mu})$\} is infeasible since they're high-dimensional deep neural networks. Inspired by generated adversarial imitation learning~\cite{ho2016generative}, we cluster different policies by the state distributions \{$\rho_i$(s)\} they generate, which are approximated by Monte-Carlo sampling. Here, $\rho_i(s)$ represents the state distribution of the scenario when the ego vehicle is in the $i$-th adversarial environment.

Another technical issue is that we do not know the explicit number of clusters. Traditional models in machine learning often require an explicit choice of the model capacity via hyperparameters. For example, in K-Means clustering which has been widely used for traffic simulation~\cite{erman2006traffic, rao2019interval}, the number of clusters must be selected a priori, which is not available in this paper. To address this issue, we turn to Bayesian nonparametric models that automatically infer the model complexity from the data. Specifically, we consider DP-Means, a prototypical Dirichlet process mixture model, which leads to a hard clustering algorithm similar to the K-Means objective~\cite{kulis2011revisiting}.  DP-Means has been widely used for the clustering of images, geological data, biological data, and music, etc~\cite{bachem2015coresets}. In contrast to K-Means, DP-Means clustering allows solutions with an arbitrary number of clusters. To use DP-Means, we must first find a suitable hyperparameter: $\lambda$, which indicates the approximate distance between different clusters. We look for it with a heuristic method suggested in~\cite{kulis2011revisiting}: Given an approximate number of desired clusters $k$, we first initialize a set $T$ with the mean distribution of \{$\rho_i$(s)\}. Then, iteratively add the distribution to $T$ which has the maximum distance to $T$. Repeat this $k$ times and set $\lambda$ as the maximum distance in the last round. To calculate the distance of two distributions, Jensen-Shannon divergence is used:

\DeclarePairedDelimiterX{\infdivx}[2]{(}{)}{%
  #1\;\delimsize\|\;#2%
}
\newcommand{\infdiv}{D_{}\infdivx}
\newcommand{\infdivjsd}{JSD_{}\infdivx}
\[
\infdivjsd{P}{Q} = \frac{1}{2}\infdiv{P}{M} + \frac{1}{2}\infdiv{Q}{M},
\]
where $M = \frac{1}{2}\left(P+Q\right)$, and $\infdiv{\cdot}{\cdot}$ is the KL divergence:
\[
\infdiv{P}{Q}=\int_s p\left(s\right)\log \frac{p\left(s\right)}{q\left(s\right)} ds.
\]
Then, with $\lambda$ and distance function, we can use DP-Means to cluster the adversarial environments. The whole algorithm of environment clustering is shown in Algorithm~\ref{clustering}. After the clustering procedure, adversarial patterns for the tested vehicle can be extracted.

\begin{algorithm}[t]

\caption{Adversarial environments clustering}
\begin{algorithmic}[1]
\label{clustering}
\STATE Run Monte-Carlo sampling to estimate \{$\rho_i$(s)\}
\STATE Heuristically set $\lambda$
\STATE Initialize $k=1$, $l_1=\{\rho_1,\dots,\rho_n\}$, global mean $\mu_1$
\STATE Initialize cluster indicators $z_i=1$ for $i=1,\dots,n$
\WHILE{\textbf{not} converge}
\FOR{$\rho_i$}
\STATE Compute $d_{ic} = \infdivjsd{\rho_i}{\mu_c}$ for $c=1,\dots,k$
\IF{$\min_c d_{ic} > \lambda$}
\STATE $k=k+1$, $z_i=k$, $\mu_k=x_i$
\ELSE
\STATE $z_i = \arg\min_c d_{ic}$
\ENDIF
\STATE Generate clusters $l_1,\dots, l_k$: $l_j=\{\rho_i|z_i=j\}$
\STATE Compute $\mu_j = \frac{1}{\abs{l_j}} \Sigma_{\rho \in l_j}\rho$
\ENDFOR
\ENDWHILE
\end{algorithmic}
\end{algorithm}

\section{Simulation}
\label{section4}
\begin{figure}[t]
\centering
\includegraphics[width=3 in]{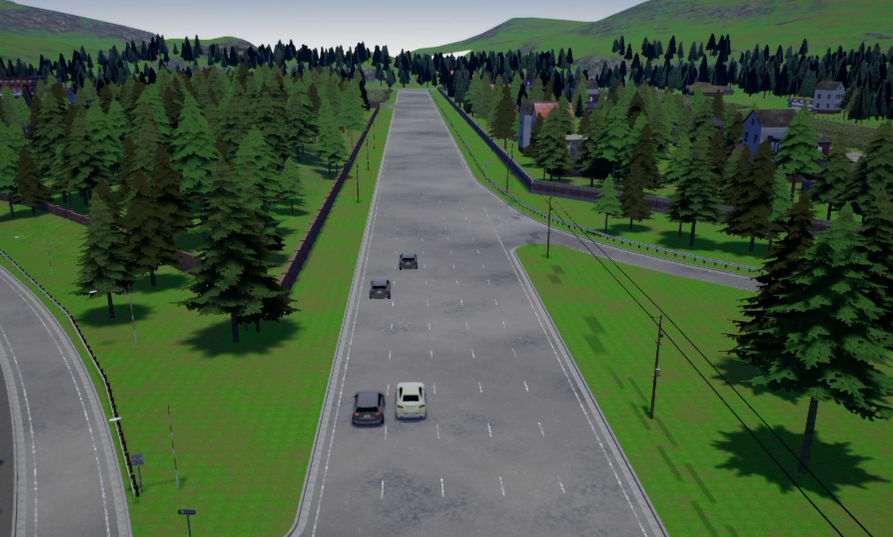}
\caption{The lane-change scenario built in the CARLA Simulator. The ego vehicle is trying to make a left lane change.}
\label{carla}
\end{figure}

The simulation in this paper is conducted with CARLA~\cite{dosovitskiy2017carla}. CARLA is an open-source simulator for autonomous driving research. It provides realistic vehicle dynamics models and supports flexible sensor configuration and scenario generation. The lane-change scenario we build in CARLA is shown in Fig.~\ref{carla}, where there are four vehicles in total: one ego vehicle that is being tested, and three adversarial vehicles namely follow vehicle, leader vehicle, and target vehicle. We use this simulation environment to demonstrate the effectiveness of the proposed adversarial evaluation approach.


\subsection{Ego Vehicle Lane-Change Controllers}
\label{sub_ego}



Two typical lane-change models are developed in this paper for evaluation. One is a traditional rule-based model using gap acceptance concepts, the other is trained by reinforcement learning. We denote the two lane-change models as $\mathcal{M}_{gap}$ and $\mathcal{M}_{rl}$, respectively.
\subsubsection{Gap Acceptance Model.} Gap acceptance is an important concept in most lane change models~\cite{ahmed1996models, toledo2003modeling}. Before executing a lane change, the driver assesses the positions and speeds of the target vehicle and the follow vehicle in the target lane (see Fig.~\ref{lanechange}) and decides whether the gap between them is sufficient for the lane change behavior. The front spacing between the ego vehicle and the leader vehicle is also critical for avoiding a front crash. The gap acceptance lane change model is developed based on~\cite{toledo2003modeling} and the SPMD database.
\subsubsection{Reinforcement Learning Model.} Reinforcement learning has become a powerful tool for the development of autonomous vehicles. However, end-to-end reinforcement learning can take a relatively long time to converge to the optimal policy. For autonomous driving, it is a good approach to develop a hierarchical framework~\cite{muller2018driving}, where reinforcement learning is only used in the high-level decision-making part, while the motion planning and control part is developed by hard-code. Specifically, in this paper, deep Q-learning~\cite{mnih2015human} is used in the decision-making part to decide whether or not to start a lane change. The desired lane change trajectory is generated by an optimal lattice planner~\cite{werling2010optimal}, while the longitudinal and lateral control of the vehicle is achieved by a model predictive controller~\cite{garcia1989model}. 

\begin{figure}[t]
\centering
\includegraphics[width=\linewidth]{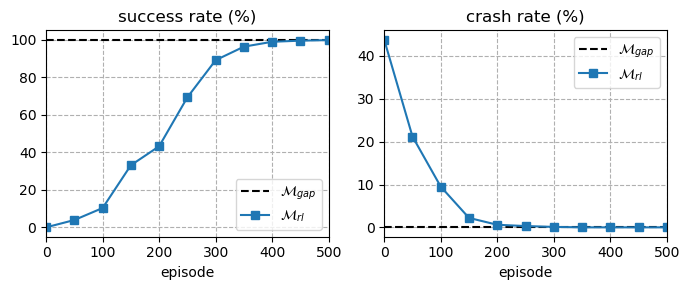}
\caption{Success rate and crash rate during training process.}
\label{fig:mrl}
\end{figure}

The success rate and crash rate during the training process is shown in Fig~\ref{fig:mrl}. Since $\mathcal{M}_{gap}$ is a rule-based model, it has a static performance. After 500 episode training, the $\mathcal{M}_{rl}$ also converges to a stable model. Both models achieved a success rate higher than 99\% in the naturalistic environment, which makes them reliable lane-change models. However, it also indicates that directly evaluating them in these naturalistic environments will be low-efficient. Next, we will generate adversarial scenarios for them and test their performances in those challenging environments.

\subsection{Adversarial MDP Setting}
As mentioned in Section~\ref{section3}, DDPG is used to train adversarial environments for the ego vehicle. We first introduce the settings of the lane-change MDP and the DDPG agent. 

The state space $S$ of the MDP is a 9-dimension vector space: [$x_{leader}$, $x_{follow}$, $x_{target}$, $v_{leader}$, $v_{follow}$, $v_{target}$, $v_{ego}$, $\phi_{ego}$, $y_{ego}$], where $x$ denotes the distance between the adversarial vehicle and ego vehicle, $v$ denotes velocity of the vehicle, $\phi_{ego}$ denotes the yaw angle of the ego vehicle, and $y_{ego}$ denotes the lateral position of the ego vehicle. Since it is a near-field scenario, we assume that all participants have perfect observation of the state information.

The adversarial vehicles only have longitudinal movements, so the action space $A$ of the adversarial agent is a 3-dimension vector space that decides their longitudinal control actions: [$a_{leader}$, $a_{follow}$, $a_{target}$]. $a$ is a float number in the range $[-1, 1]$, where $+1$ indicates a full-throttle and $-1$ indicates a full brake. Since DDPG can already handle continuous action space, no discretizations are needed.

The reward function of adversaries is $r_{adv} = -r_{ego} + \beta r_{rule}$, as introduced in Section~\ref{adv}. Here, $r_{ego}$ is the reward function of the ego vehicle and should indicate the driving performance of it. In the lane-change scenario, 
\[
  r_{ego} =
  \begin{cases}
                   100 & \text{if a lane-change is finished} \\
                   -50 & \text{if a collision happened} \\
                   0.1v_{ego} & \text{otherwise} \\
  \end{cases},
\]
which awards the agent for completing the lane-change task, and penalizes it for collisions. The velocity term provides a dense reward to encourage the ego vehicle to go faster, which is commonly used in related papers~\cite{lillicrap2015continuous}. For the adversarial vehicles, $r_{rule} = -50$ if the adversarial vehicles break the traffic law, otherwise $r_{rule} = 0$. We set $\beta=1$ as the default value and will show the influence of it in the later section.



\begin{table}[t]
\renewcommand{\arraystretch}{1.3}
\caption{Hyperparameters for training}
\label{tab:hyp}
\centering
\begin{tabular}{c c}
\hline
Parameter & Value\\
\hline\hline
discount factor $\gamma$ & 0.99 \\
actor learning rate & 0.005 \\
critic learning rate & 0.01 \\
soft update rate & 0.01 \\
batch size & 128 \\
buffer size & 10000 \\
ensemble size & 100 \\
\hline
\end{tabular}
\end{table}

\subsection{Results of Adversarial Training}

With Algorithm~\ref{ensemble} and~\ref{clustering}, We train adversarial agents for $\mathcal{M}_{gap}$ and $\mathcal{M}_{rl}$. The hyperparameters used for training is shown in Tab.~\ref{tab:hyp}, which are decided by a hyperparameter search. The actor model used in DDPG is a three-layer fully-connected neural network with the number of hidden units: [64, 64, 3]. The activation function is ReLU for the first two layers and Tanh for the output layer to get a [-1, 1] output. The critic model is a four-layer fully-connected neural network with the number of hidden units: [64, 64, 32, 1]. The activation is ReLU for the first three layers and Identity for the output layer. The ensemble size is 100 for each lane-change model, which means we train 100 adversarial agents against $\mathcal{M}_{gap}$ and $\mathcal{M}_{rl}$, respectively.

\begin{figure}[b]
\centering
\includegraphics[width=\linewidth]{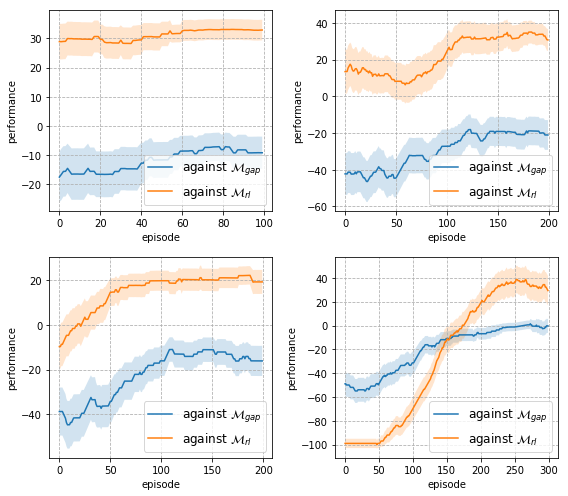}
\caption{Learning curves of adversarial agents against different lane-change models.}
\label{fig:learning}
\end{figure}

We show 4 sets of learning curves in Fig.~\ref{fig:learning}, where the performance represents the episodic accumulative reward achieved by the adversarial agents. According to the figure, each bootstrap adversarial policy in the ensemble has converged to a stable model when stops training, which indicates that they have successfully found local optimums as planned. Also, different performance at convergence indicates that adversaries have learned different ways to challenge the ego vehicle. We will cluster them and illustrate the generated risky scenarios in the later section.

\begin{figure}[t]
\centering
\includegraphics[width=\linewidth]{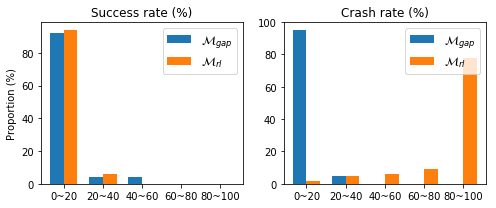}
\caption{Histograms of success rate and crash rate of tested lane-change models in adversarial environments.}
\label{fig:hist}
\end{figure}

An interesting observation from Fig.~\ref{fig:learning} is that the performance of adversaries against the rule-based model $\mathcal{M}_{gap}$ is lower than those against the learning-based model $\mathcal{M}_{rl}$. In other words, the $\mathcal{M}_{gap}$ developed in this paper shows to be more robust to adversarial challenges compared to the $\mathcal{M}_{rl}$. To better illustrate this, we run Monte-Carlo simulation with all learned adversarial policies and draw histograms of the success rate and the crash rate in Fig.~\ref{fig:hist}. It is clearly shown that both lane-change models have a low success rate in adversarial environments. However, based on the crash rate, $\mathcal{M}_{gap}$ turns out to be a much safer controller than $\mathcal{M}_{rl}$. The results indicate that in most runs in the adversarial scenarios, the $\mathcal{M}_{gap}$ got stuck by the adversarial vehicles, while the $\mathcal{M}_{rl}$ failed to keep safe and experienced a crash accident. The results are not surprising: the rule-based model $\mathcal{M}_{gap}$ is build based on human knowledge, so it is more robust to unseen environments. On the contrary, the learning-based model $\mathcal{M}_{rl}$ is trained only in naturalistic scenarios, so it has difficulty generalizing to more risky ones generated by adversarial agents.

\begin{figure}[h]
\centering
\includegraphics[width=\linewidth]{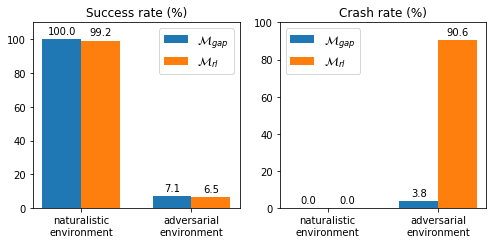}
\caption{Success rate and crash rate of $\mathcal{M}_{gap}$ and $\mathcal{M}_{rl}$ in naturalistic and adversarial environments.}
\label{fig:overall}
\end{figure}

Finally, we compare the overall performance of $\mathcal{M}_{gap}$ and $\mathcal{M}_{rl}$ in naturalistic and adversarial environments in Fig.~\ref{fig:overall}. It is clearly shown that the generated adversarial environments significantly degrade the performance of both $\mathcal{M}_{gap}$ and $\mathcal{M}_{rl}$. Especially for $\mathcal{M}_{rl}$, the performance of 99.2\% success rate in naturalistic environments turns to 90.6\% crash rate in adversarial environments, which is a catastrophic transition. 

\begin{figure}[t]
     \centering
     \begin{subfigure}[b]{0.6\linewidth}
         \centering
         \includegraphics[width=\textwidth]{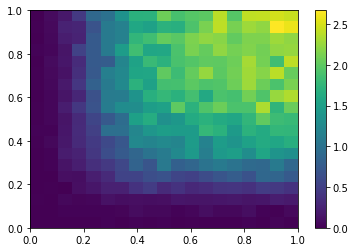}
         \caption{Naturalistic state distribution.}
         \label{fig:nat_hist}
     \end{subfigure}
     \hfill
     \begin{subfigure}[b]{0.4\linewidth}
         \centering
         \includegraphics[width=\textwidth]{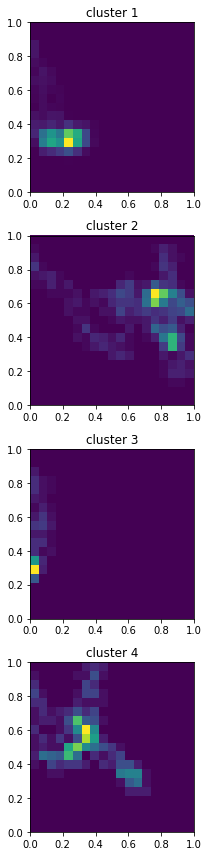}
         \caption{Adversarial distributions for $\mathcal{M}_{gap}$.}
         \label{fig:gap_clus}
     \end{subfigure}
     \begin{subfigure}[b]{0.4\linewidth}
         \centering
         \includegraphics[width=\textwidth]{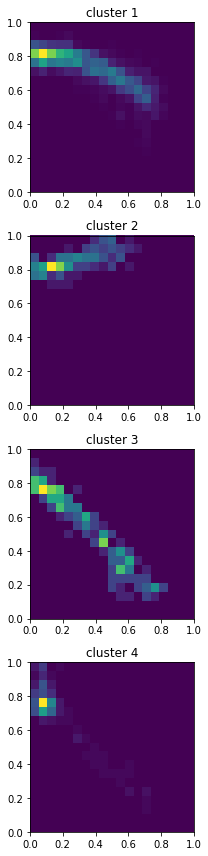}
         \caption{Adversarial distributions for $\mathcal{M}_{rl}$.}
         \label{fig:rl_clus}
     \end{subfigure}
        \caption{State distributions of naturalistic scenarios and generated adversarial scenarios.}
        \label{fig:clus}
\end{figure}



\begin{figure*}[t]
     \centering
     \begin{subfigure}[b]{0.9\linewidth}
         \centering
         \includegraphics[width=\textwidth]{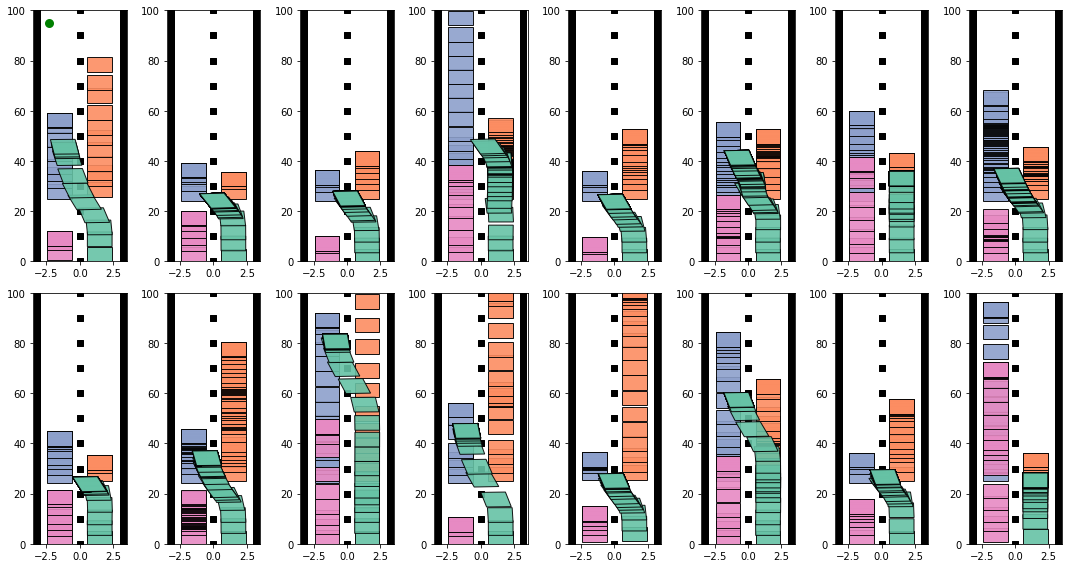}
         \caption{Generated adversarial scenarios for $\mathcal{M}_{gap}$.}
         \label{fig:gaps}
     \end{subfigure}
     \hfill
     \begin{subfigure}[b]{0.9\linewidth}
         \centering
         \includegraphics[width=\textwidth]{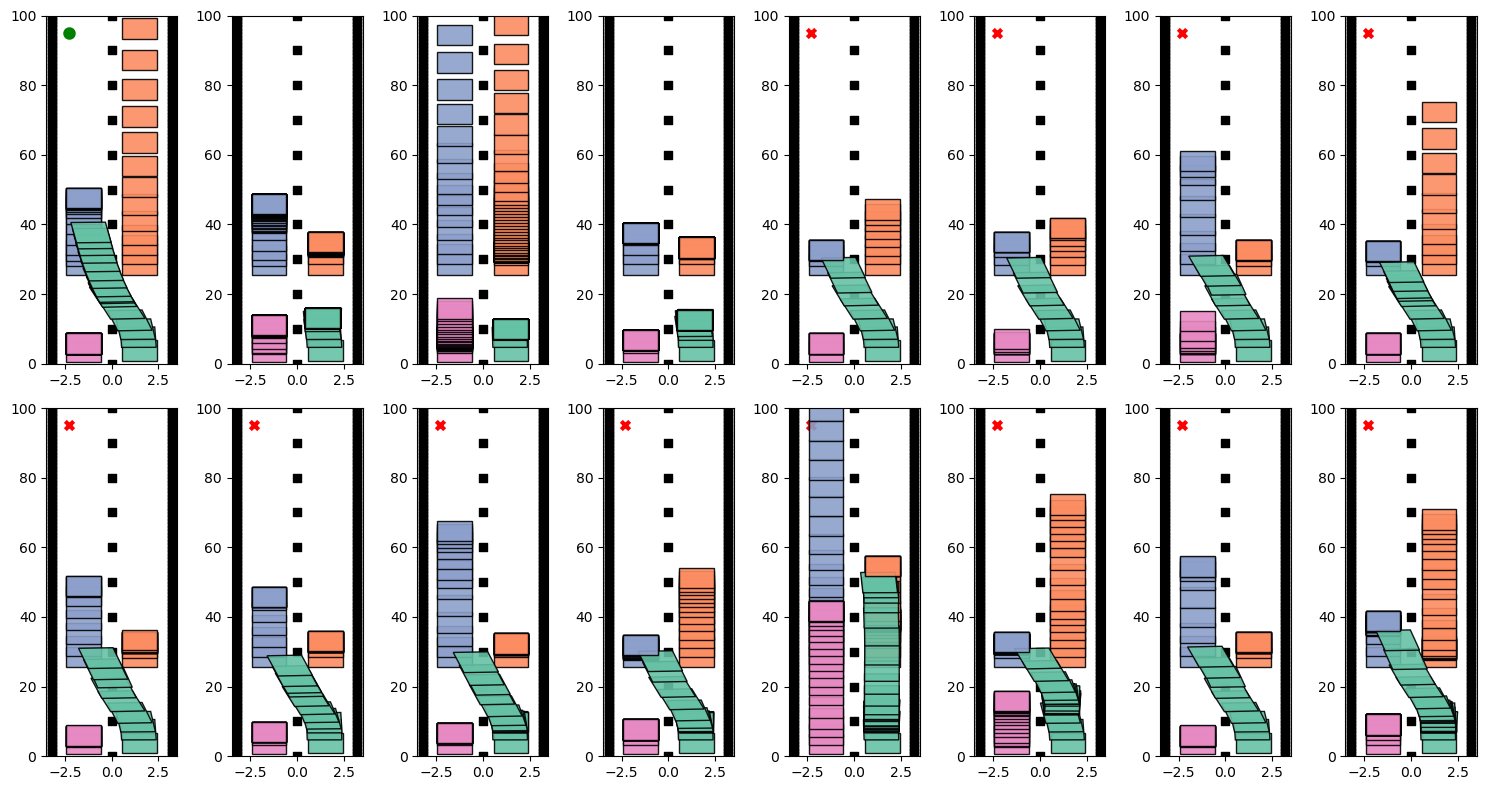}
         \caption{Generated adversarial scenarios for $\mathcal{M}_{rl}$.}
         \label{fig:rls}
     \end{subfigure}
        \caption{Generated adversarial scenarios.}
        \label{fig:scens}
\end{figure*}


\subsection{Clustering and Visualization}
In this section, we will focus on the clustering and visualization of the statistical results presented in the previous section.
Since the original state space is 9-dimensional, we reduce the dimension with principal component analysis~\cite{wold1987principal} for visualization. 

The reduced 2-dimensional state distributions of the naturalistic scenarios and the generated adversarial scenarios are shown in Fig~\ref{fig:clus}. Based on these histograms, the generated adversarial distributions (Fig.~\ref{fig:gap_clus} ,\ref{fig:rl_clus}) are much more narrow than the naturalistic distribution (Fig.~\ref{fig:nat_hist}). Moreover, in most cases, the adversarial distributions concentrate on states that are rare for the naturalistic distribution. Given the statistical results of the performance, the adversarial distributions also represent more risky scenarios compared to the naturalistic one. Thus, these figures clearly indicate that our approach can efficiently find rare risky scenarios for the evaluation of autonomous vehicles. In this sense, our method can be regarded as an efficient searching algorithm in the high-dimensional policy space.

Next, we visualize the generated adversarial patterns for $\mathcal{M}_{gap}$ and $\mathcal{M}_{rl}$ in Fig.~\ref{fig:scens}. For $\mathcal{M}_{gap}$, the adversaries in most patterns learn to prevent a successful lane change by blocking in front of the ego vehicle, as shown in Fig.~\ref{fig:gaps}. The blocking is usually done by the leader vehicle and the target vehicle, while the follow vehicle is trying to minimize the lane-change gap. With these challenges, the rule-based lane-change model $\mathcal{M}_{gap}$ is able to avoid crashes in most experiments, but the success rate of lane change is low at \%7.1, as shown in Fig.~\ref{fig:overall}.



For $\mathcal{M}_{rl}$, the adversarial scenarios are shown in Fig.~\ref{fig:rls}. The ego vehicle's behavior is sometimes too aggressive to collide with the target vehicle or the leader vehicle. It can also be too conservative to stop with safe lane-change space. The inconsistent and unpredictable behavior of the ego vehicle can be explained by the state distribution mismatch between the generated adversarial environments and the training environments shown in Fig.~\ref{fig:clus}: the ego vehicle can not deal with adversarial scenarios that never appear during training. Thus, the adversarial agents can easily exploit the policy of $\mathcal{M}_{rl}$, leading to a 90.6\% crash rate. 

\subsection{Influence of the Reward Configuration}

In this section, we discuss about the influence of the hyperparameter $\beta$, which plays an important role in the configuration of the adversarial reward function: $r_{adv} = -r_{ego} + \beta r_{rule}$, where $r_{ego}$ represents the driving performance of the ego vehicle, and $r_{rule}$ is a penalty for violation of traffic rules. With this penalty, the adversarial vehicles in the environment will try to obey the traffic rules when interfering with the ego vehicle.

\begin{figure}[h]
\centering
\includegraphics[width=0.6\linewidth]{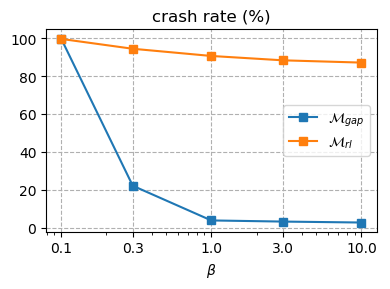}
\caption{Crash rate of $\mathcal{M}_{gap}$ and $\mathcal{M}_{rl}$ with different $\beta$.}
\label{fig:beta}
\end{figure}

\begin{figure}[h]
\centering
\includegraphics[width=0.5\linewidth]{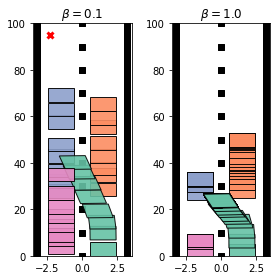}
\caption{A comparison of typical risky scenarios generated with different $\beta$.}
\label{fig:comp}
\end{figure}

The statistical results of the crash rate as $\beta$ changes are shown in Fig.~\ref{fig:beta}. According to the results, the crash rate drops as $\beta$ increases and is significantly high when $\beta$ is a small value (0.1). This is a reasonable result because $\beta$ represents the rationality of the adversarial vehicles: with a small value of $\beta$, the adversarial agents will try everything they can to make a crash, without considering traffic laws. To better illustrate this, we compare the typical scenarios with different $\beta$ in Fig.~\ref{fig:comp}. When $\beta=0.1$, the follow vehicle (pink) learns to suddenly accelerate and rush to the ego vehicle to make a crash, even though it will be the vehicle in charge of this accident. Thus, we can control the rationality of the environment vehicles by tuning $\beta$. In this paper, we set $\beta=1$ because we found it is able to generate diverse adversarial behaviors that are neither too aggressive nor too conservative. However, a more thorough study of the reward configuration is potentially valuable.


\section{Conclusion}
\label{section5}
In this paper, we propose an adversarial framework for the efficient evaluation of autonomous vehicles. Adversarial risky scenarios are generated with ensemble deep reinforcement learning and clustered with a nonparametric Bayesian method. The simulation results show that the adversarial environments can significantly degrade the performance for both rule-based and learning-based lane-change models. Thus, this approach can be regarded as a promising way to supplement the current evaluation framework of autonomous vehicles.

A promising future direction is to improve the autonomous driving strategy guided by the extracted weaknesses from the adversarial evaluation. More realistic assumptions, such as partially-observable states, loss of observation signals, and human emotions, can also be incorporated to enrich this framework.

\section{Acknowledgement}

This paper was supported in part by the National Key Research and Development Program of China under Grant 2018YFB0105101 and in part by Technological Innovation Project for New Energy and Intelligent Networked Automobile Industry of Anhui Province under Grant JAC2019030105.


%





\ifCLASSOPTIONcaptionsoff
  \newpage
\fi



\bibliographystyle{IEEEtran}
\bibliography{IEEEabrv,bib}
\end{document}